\documentclass[11pt,a4paper]{article}
\usepackage[T1]{fontenc}
\usepackage{times}
\usepackage[a4paper,margin=24mm,top=22mm,bottom=23mm]{geometry}
\usepackage{amsmath,amssymb,graphicx,booktabs}
\usepackage{microtype}
\usepackage[numbers,sort&compress]{natbib}
\usepackage[hyphens]{url}
\usepackage{placeins,needspace}
\usepackage{enumitem}
\usepackage[font=small,labelfont=bf,labelsep=period,skip=6pt]{caption}
\usepackage[hidelinks,breaklinks]{hyperref}
\title{Where to Look and at What Precision:\\Action-Conditioned Visual Allocation for Robot Manipulation}
\author{Zhenglong Qian\qquad Zixi Jia\qquad Tianyi Zhang}
\date{}
\hypersetup{
  pdftitle={Where to Look and at What Precision: Action-Conditioned Visual Allocation for Robot Manipulation},
  pdfauthor={Zhenglong Qian, Zixi Jia, Tianyi Zhang},
  pdfsubject={Preprint},
  pdfkeywords={robot manipulation, visual precision, action-conditioned routing, ACT, LIBERO, RLBench}
}
\makeatletter
\def\@maketitle{%
  \newpage\null\vskip 2pt
  \begin{center}
    {\fontsize{16}{20}\selectfont\bfseries\@title\par}
    \vskip 12pt
    {\large\normalfont\@author\par}
    \vskip 14pt
  \end{center}
}
\renewenvironment{abstract}{%
  \begin{center}\bfseries Abstract\end{center}
  \begin{quote}\small\noindent\ignorespaces
}{\end{quote}\vspace{5pt}}
\makeatother
\setlist[itemize]{itemsep=3pt,topsep=4pt}

\begin{document}
\maketitle
\begin{abstract}
Identifying a task-relevant region does not determine the visual precision or additional capacity it requires. We recast the global resolution trade-off as local visual configuration beyond a fixed base observation. With the same local input size, denser sampling preserves finer details but reduces spatial coverage. Maintaining coverage requires more observations. We hold the Base policy and candidate precision levels fixed to examine how visual supplementation and additional capacity affect control. Among the tested configurations, control gains do not increase monotonically with the number of observations, and the preferred capacity varies across tasks. Supplementary value must be evaluated through action outcomes, but encoding all candidates before selection incurs the full processing cost. The number of candidates varies across precision levels, yet all candidates must be compared under a shared budget. We propose action--visual precision matching, which uses the base policy's action and visual context to score candidates before additional encoding. We train this allocation by comparing action errors between supplementary sets. After cardinality correction, location--precision candidates compete in a global Top-$K$. Only selected local observations are encoded, and Base-aligned visual residuals correct the actions. We combine action predictions from different prefixes of the same ranking through learned multi-capacity fusion, and evaluate the complete system against Base policies with different input resolutions and training budgets on LIBERO and RLBench10.
\end{abstract}
\section{Introduction}
The location of a task-relevant region alone does not determine the required visual precision. Within the same region, different details may play different roles in the current action. Visual relevance must therefore be considered in relation to visual precision and visual capacity.

Existing vision-language-action models typically use fixed-resolution inputs to balance control performance and computational cost~\cite{kim2024,black2024pi0}. The input specifications and encoding scheme remain fixed, while image content varies with the task and state. Prior work improves the use of information already present in the base input through compression, retention, resampling, or attention-based selection~\cite{rao2021dynamicvit,ryoo2021,bolya2023,gao2025compressor,pei2025adp}. These operations cannot supply original details that were lost during downsampling. Globally increasing input resolution provides access to more detail, but also increases the amount of data processed and the computational cost. In OpenVLA, increasing input resolution from $224\times224$ to $384\times384$ approximately tripled training time without improving control performance~\cite{kim2024}. This result indicates that the benefits of increasing global visual resources should be evaluated in relation to control requirements.

We recast the global resolution trade-off as a local observation configuration problem beyond a fixed Base. With a fixed number of sampled pixels per local input, higher relative visual precision requires a smaller covered area in the original image. Maintaining the same spatial coverage then requires more observations and additional tokens, creating a trade-off between precision, coverage, and capacity. Under a limited additional-token budget, the choice between broader spatial relations and finer local detail depends on what each observation adds to the current action beyond Base. We fix the candidate precision levels and control visual capacity through the number of added tokens. Within each budget, we learn joint location--precision allocation and examine how capacity configurations affect control.

We evaluate a candidate's supplementary value through its effect on action generation. Encoding all high-precision candidates before selection would still incur the full processing cost. Their coverage and number also vary across precision levels, requiring comparison under a shared budget. We propose action--visual precision matching. It uses Base's existing hidden action representation and visual context to score candidates before additional encoding, and learns their supplementary value from action prediction errors. All location--precision candidates compete in one global ranking. The selected count $K$ controls additional capacity, and only selected observations are encoded to correct Base actions. Base remains frozen during additional training; only the added branch is updated.

Using ACT~\cite{zhao2023act}, we evaluate local visual supplementation on LIBERO~\cite{liu2023libero} and RLBench10~\cite{james2020rlbench}. We compare the effects of additional capacity, base input resolution, and training budget on closed-loop performance, and analyze computational cost. LIBERO results show that control gains do not increase monotonically with additional count and that preferred capacities differ across tasks. Local supplementation can outperform higher-resolution Base policies trained for longer. We further combine action predictions from different capacities under the same candidate ranking and evaluate multi-capacity action fusion in closed-loop control.

\Needspace{4\baselineskip}
Our contributions are as follows:
\begin{itemize}
\item \textbf{Problem formulation.} We study local visual supplementation in terms of the additional action-related information it provides beyond a fixed base policy. We examine the trade-off between visual precision, spatial coverage, and additional visual capacity.
\item \textbf{Mechanism.} We develop action-conditioned location--precision allocation with a single global Top-$K$ selection before additional visual encoding. The allocation is learned from comparisons of action prediction errors, and the selected observations refine Base actions through aligned visual residuals. We further combine action predictions from different prefixes of the same candidate ranking.
\item \textbf{Evaluation.} Using ACT on LIBERO and RLBench10, we compare local supplementation with increasing the base input resolution and extending the training budget. We analyze how control performance varies with added capacity and task, and evaluate multi-capacity action fusion together with its computational cost.
\end{itemize}

\section{Related work}
\subsection{Using visual information already present in the base input}
Visual token pruning, aggregation, and merging reduce redundancy in existing visual representations~\cite{rao2021dynamicvit,ryoo2021,bolya2023}. Multi-scale feature hierarchies preserve representations at different spatial scales~\cite{lin2017}. Focus-DETR selects informative tokens using localization and category information~\cite{zheng2023focus}. For vision-language-action models, Compressor-VLA uses language instructions to guide visual token compression while retaining task-relevant semantic and spatial information~\cite{gao2025compressor}. ADP uses action history to adjust visual token retention across manipulation stages~\cite{pei2025adp}.

Other work preserves and uses local spatial information. Patch Policy directly consumes dense features from a pretrained visual encoder, reducing information loss from global aggregation~\cite{zhou2026patchpolicy}. AVA-VLA combines language instructions and execution history to adjust the attention weights of current visual tokens~\cite{xiao2025ava}. These methods primarily improve the use of existing visual information. We focus on supplementary local observations beyond the base input.

\subsection{Local visual refinement and supplementary observations}
Recurrent visual attention learns task-dependent glimpse locations and processes local observations at multiple scales~\cite{mnih2014recurrent}. Attention sampling and differentiable patch selection process selected image regions to reduce the cost of high-resolution inputs~\cite{katharopoulos2019sampling,cordonnier2021patch}. In robot control, Multi-Resolution Sensing combines spatial and temporal resolutions, using low-frequency global information and high-frequency local feedback for precise control~\cite{saxena2023multires}. AtVLA uses action-conditioned attention to locate local regions when action predictions are uncertain. It re-encodes the crops and appends them to the existing context to refine action generation~\cite{cui2026atvla}. ActiveVLA combines critical-region localization in 3D, virtual viewpoint selection, and local zooming to acquire observations for manipulation~\cite{liu2026activevla}.

Side-Tuning adapts a frozen backbone with an added trainable network~\cite{zhang2020sidetuning}. Fast-in-Slow uses latent context from its reasoning system to condition action generation~\cite{chen2025fastinslow}. Here, Base supplies the existing action context used to read selected visual residuals.

\begin{figure}[!htbp]
\centering
\includegraphics[width=0.84\linewidth,height=0.30\textheight,keepaspectratio]{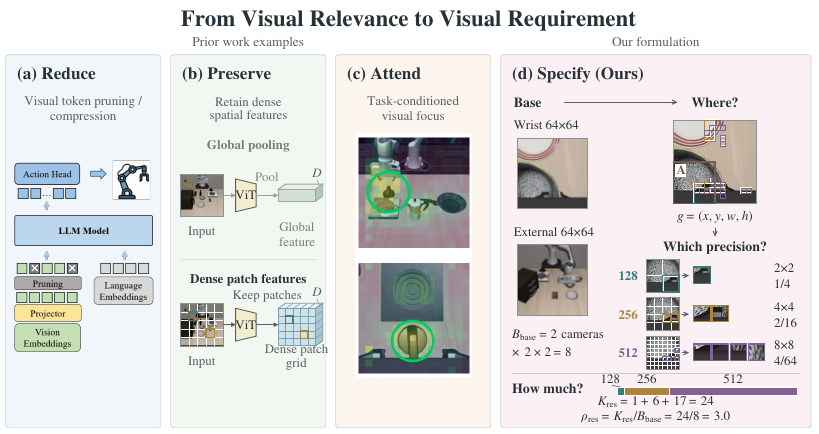}
\caption{Visual information and supplementary configuration. (a) Token-pruning pipeline adapted from Gao et al.~\cite{gao2025compressor}, \href{https://arxiv.org/html/2511.18950v1}{Fig.~1(b)}, with added deletion marks. (b) Our conceptual illustration of global pooling versus dense patch features based on Patch Policy~\cite{zhou2026patchpolicy}; the same observation image, cropped from \href{https://arxiv.org/html/2607.18236v1}{its Fig.~2}, is reused in both rows. (c) Attention visualizations cropped from AVA-VLA~\cite{xiao2025ava}, \href{https://arxiv.org/html/2511.18960v1}{Fig.~1(b)}. Source material in (a)--(c) is reused under \href{https://creativecommons.org/licenses/by/4.0/}{CC BY 4.0}; excerpts are cropped, resized, and rearranged. (d) Our configuration jointly specifies location, visual precision, and additional count.}
\label{fig:overview}
\end{figure}

We study local visual supplementation beyond a fixed Base observation. In Fig.~\ref{fig:overview}(d), Base provides action and visual context for assessing supplementary needs. The supplementary configuration specifies location, relative visual precision, and additional count. We fix the candidate precision levels, control capacity through $K$, and jointly allocate locations and precision levels within each budget. We compare control performance across capacities.

\FloatBarrier
\section{Method}

\subsection{Base-aligned visual supplementation and action correction}

\noindent\textbf{Base observations and supplementary inputs.} Base observations and supplementary inputs come from native \(512\times512\) images captured by the same camera at the same time. Successive downsampling produces the \(512,256,128,64\) image levels, which share the same scene state and action supervision. Base uses \(64\times64\) images from the external and wrist cameras. Supplementary observations come from higher-resolution levels. They provide real details retained in the original images, rather than content generated by enlarging Base images. Section 4 reports the comparisons of base resolutions.

Let the base context be \(B_t=(I_t^{64},s_t)\), where \(I_t^{64}\) contains the two camera inputs and \(s_t\) is the robot state. ACT~\cite{zhao2023act}, with a ResNet18 backbone~\cite{he2016resnet}, produces the hidden action representation \(H_t=f_B(B_t)\). Its native output layer then predicts \(\hat{\mathbf a}_t^B=g_B(H_t)\). Here, \(H_t\in\mathbb R^{50\times512}\) precedes the action output mapping and represents the next 50 action steps. Base remains frozen during additional training. The same \(H_t\) supports native action prediction and conditions the readout of supplementary visual information.

\begin{figure}[!htbp]
\centering
\includegraphics[width=1.0\linewidth,height=0.36\textheight,keepaspectratio]{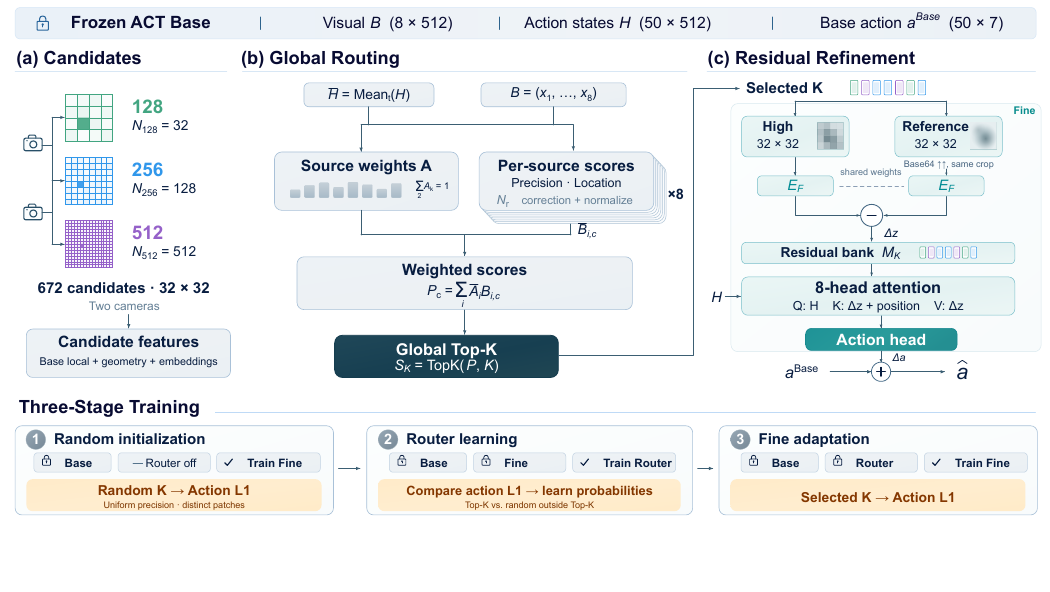}
\caption{Action--visual precision matching. (a) Candidate descriptors from Base information and geometry. (b) Per-source scoring, cardinality correction, weighted mixing, and global Top-$K$. (c) Selected High--Reference residuals refine actions through eight-head attention. Training uses Random Fine1E, fixed-Fine Router1E, and Top-only Fine2E; Base stays frozen.}
\label{fig:method}
\end{figure}
\FloatBarrier
\noindent\textbf{Relative visual precision and additional action information.} Local observations come from the \(128,256,512\) image levels, each with an input size of \(32\times32\). For candidate \(c\), let \(n\) be the sampled pixel count and \(a_c\) its covered area in the original image. Visual precision and the relative precision of two candidates are defined as:

\begin{equation}
\rho(c)=\frac{n}{a_c},\qquad \frac{\rho(c_h)}{\rho(c_l)}=\frac{a_l}{a_h}.\label{eq:1}
\end{equation}

With equal sampled pixel counts, a smaller covered area gives higher visual precision. Patches from the three levels cover \(128\times128,64\times64,32\times32\) regions in the original image, respectively. Their visual precision ratio is \(1:4:16\). This measures image sampling, not the robot's physical positioning accuracy.

Equal input pixel counts do not imply equally useful action information. Let \(\mathbf a_t^*\) be the target action sequence and \(X_c\) a candidate local observation. We define its additional action information beyond Base using conditional mutual information~\cite{cover2006elements}:

\begin{equation}
\mathcal R(c\mid B_t)=I(\mathbf a_t^*;X_c\mid B_t).\label{eq:2}
\end{equation}

This quantity describes the average action information gained beyond what Base already provides. Consider a higher-precision observation \(X_h\) and a lower-precision observation \(X_l\). When the relevant mutual information quantities are finite, the chain rule~\cite{cover2006elements} gives:

\begin{equation}
\begin{aligned}
&\mathcal R(c_h\mid B_t)-\mathcal R(c_l\mid B_t)\\
&\quad=I(\mathbf a_t^*;X_h\mid B_t,X_l)\\
&\qquad-I(\mathbf a_t^*;X_l\mid B_t,X_h).
\end{aligned}\label{eq:3}
\end{equation}

Each term on the right measures what one observation adds beyond Base and the other observation. A positive difference means that the higher-precision observation provides more additional action information. Thus, whether finer local detail is more useful than broader coverage depends on what it adds for the target action, not on visual precision alone. Conditional mutual information defines this comparison. Our method learns to use supplementary observations through action prediction error, without directly estimating this information quantity.

\noindent\textbf{Base-aligned visual residuals.} We represent supplementary information by comparing features from the real observation and the Base reference for the same region. Let \(\mathcal S_t\) denote the supplementary set; later subsections define its size and selection. For each candidate \(c\), we read the real patch \(X_c^{\mathrm{High}}\) from its precision level. We bilinearly upsample the same camera's Base64 image to that level and crop at the same location to obtain \(X_c^{\mathrm{Reference}}\). Both patches correspond to the same time, camera, and original-image region, and use the same encoding input size.

High and Reference share the local encoder \(E_F\). Their encoded difference forms the visual residual:

\begin{equation}
\Delta z_c=E_F(X_c^{\mathrm{High}})-E_F(X_c^{\mathrm{Reference}}).\label{eq:4}
\end{equation}

Relative precision determines which local observation is read. The feature difference represents how that observation differs from its Base-aligned reference and supplies additional input for action generation.

\noindent\textbf{Action-conditioned readout and correction.} To use supplementary information for the current action, Fine generates queries from \(H_t\). It forms both keys and values from visual residuals, and adds candidate geometric position encodings to the keys:

\begin{equation}
\begin{aligned} Q&=H_tW_Q,\\ K_c&=\Delta z_cW_K+\operatorname{PE}(c)W_{K,\mathrm{pos}},\\ V_c&=\Delta z_cW_V. \end{aligned}\label{eq:5}
\end{equation}

Here, \(\operatorname{PE}(c)\) is the candidate position encoding. Queries retain the full 50-step action context. The visual content of both keys and values comes from the High--Reference difference.

The selected candidates form the keys \(K_{\mathcal S_t}\) and values \(V_{\mathcal S_t}\). Eight-head attention~\cite{vaswani2017} reads the supplementary information, with 64 dimensions per head. The head outputs are concatenated and passed through a projection without bias. A single 50-step correction head \(g_F\) converts this readout into an additive action residual:

\begin{equation}
\begin{aligned}
\hat{\mathbf a}_t^F&=\hat{\mathbf a}_t^B\\
&\quad+g_F\!\left(\operatorname{MHA}\left(Q,K_{\mathcal S_t},V_{\mathcal S_t}\right)\right).
\end{aligned}\label{eq:6}
\end{equation}

The correction path satisfies \(g_F(0)=0\). If High and Reference have identical encoded features for all supplementary observations, the visual residuals and attention readout are zero. The output then reduces to native Base actions.

\noindent\textbf{Action prediction objective.} We evaluate the final prediction and train Fine with L1 loss \(\mathcal L_{\mathrm{act}}\) over valid positions in the next 50 action steps. Padding at the end of a demonstration is excluded. This loss evaluates the supplementary set as a whole; it does not provide contribution labels for individual candidates.

\subsection{Additional visual capacity and supplementation under a budget}

\noindent\textbf{Candidates and capacity.} We construct non-overlapping \(32\times32\) patches on the \(128,256,512\) image levels of both cameras. These levels provide \(32,128,512\) patches, respectively, forming the 672-candidate set \(\mathcal C\). Processing the full set requires encoding every High and Reference patch. We limit this work by defining additional visual capacity as the selected patch count \(K\). Each step encodes only \(K\) High and \(K\) Reference patches, producing \(K\) residual tokens.

With the same local input size and encoder, encoding all candidates requires \(672/K\) times the local computation of encoding the selected set, or about 28 times at \(K=24\). This ratio excludes Base, candidate scoring, preprocessing, and attention readout. Section 4.4 distinguishes local encoding cost from full-network computation and measured latency.

\noindent\textbf{Selection before encoding.} Given candidate scores \(s_t(c)\), a single global Top-\(K\) operation selects the supplementary set:

\begin{equation}
\mathcal S_t=\operatorname{TopK}_{c\in\mathcal C}\bigl(s_t(c),K\bigr).\label{eq:7}
\end{equation}

Each selected index specifies the camera, image level, and crop location. We read only the corresponding High patches and construct their spatially aligned References. The supplementary path in Section 3.1 then corrects the actions. Selection precedes additional visual encoding, so unselected candidates need not be encoded. Section 3.3 introduces candidate scoring.

\noindent\textbf{Random pretraining.} Before learning candidate scores, we train Fine with random supplements balanced across precision levels. This draws on MDSR's use of randomly selected scales to organize training inputs~\cite{lim2017edsr}. For each of the \(K\) slots, we choose a precision level with equal probability. We then sample the assigned number of locations without replacement from each level's candidates across both cameras. This balances expected sample counts across precision levels. It neither fixes the proportions in a single set nor constrains later routing.

In this stage, Base is frozen and Router is not updated. Only Fine is pretrained for 1E with the action L1 loss, learning to use randomly sampled local visual residuals. Random selection does not use the current action context, and equally sized sets may produce different action errors. We use these differences to learn candidate scores and allocate capacity across locations and precision levels.

\subsection{Action-conditioned global competition across locations and precision levels}

We score candidates before additional encoding using Base's action and visual context. At a fixed additional capacity, dense sampling of small regions and broader spatial information may differ in their value for actions. Action errors after supplementation guide the allocation across locations and precision levels.

\noindent\textbf{Action-conditioned candidate scoring.} Router uses the temporal mean of the 50-step action representation \(H_t\) as its condition. This mean and each Base visual token \(x_i\) are projected into \(q\) and \(k_i\), respectively. They are then combined into a context \(z_i\). We compute the context mixture weights and precision probabilities as:

\begin{equation}
\begin{aligned}
A_i&=\operatorname{softmax}_i\left(\frac{q^\top k_i}{\sqrt{64}}\right),\\
R_{i,r}&=\operatorname{softmax}_r(W_Rz_i).
\end{aligned}\label{eq:8}
\end{equation}

The candidate descriptor $\phi_c$ in Fig.~\ref{fig:method}(a) combines existing Base visual information for the candidate region with its location, coverage, camera, and precision. The scoring module in Fig.~\ref{fig:method}(b) combines $\phi_c$ with the action-conditioned visual context $z_i$ to compute $u_{i,c}=f_C(z_i,\phi_c)$. Each context scores all 672 candidates across both cameras without first reading High features. Location scores are normalized within each precision level across both cameras:

\begin{equation}
C_{i,r,c}=\frac{\exp(u_{i,c})}{\displaystyle\sum_{c'\in\mathcal C_r}\exp(u_{i,c'})},\qquad c\in\mathcal C_r.\label{eq:9}
\end{equation}

Here, \(R_{i,r}\) evaluates precision, while \(C_{i,r,c}\) evaluates locations within that precision level. Each visual context evaluates all 672 candidates, without intermediate filtering by region or precision.

\noindent\textbf{Cardinality correction and global competition.} Precision levels contain different numbers of candidates. Because location probabilities are normalized within each level, candidate count affects the probability scale of individual candidates. We correct for this effect and normalize within each visual context, then mix the distributions using \(A_i\):

\begin{equation}
\begin{aligned} \widetilde B_{i,c}&=\frac{N_{r(c)}R_{i,r(c)}C_{i,r(c),c}}{\displaystyle\sum_rN_rR_{i,r}},\\ P_t(c)&=\sum_{i=1}^{8}A_i\widetilde B_{i,c}. \end{aligned}\label{eq:10}
\end{equation}

Here, \(N_r=|\mathcal C_r|\), and \(r(c)\) denotes the candidate's precision level. In the uniform reference case, precision levels have equal probability, as do locations within each level. Every candidate then has a final probability of \(1/672\).

Each visual context first applies cardinality correction and normalization independently. The distributions are then combined using $A_i$ to obtain the global candidate probability $P_t(c)$. These operations correspond to the independent scores, weighted aggregation, and global selection in Fig.~\ref{fig:method}(b). This order preserves the original mixture weights. Using $P_t(c)$ as the score in Eq.~\ref{eq:7}, we select $K$ candidates once and send their High and Reference patches to the residual correction path in Fig.~\ref{fig:method}(c).

\noindent\textbf{Action error feedback.} In the second stage, Base and Fine are frozen, and only Router is trained. Router training lasts 1E, measured as update opportunities. For each sample $b$, we compare the current Top-$K$ ranking $T_b$ against a disjoint random set $R_b$, evaluated by the same Fine. The random candidates are sampled uniformly without replacement from the remaining candidates. For the evaluated budget set $\mathcal K$, the action error is
\begin{equation}
D_b^q=\frac{1}{|\mathcal K|}\sum_{k\in\mathcal K}
\mathcal L_{\mathrm{act}}\!\left(S_b^q[1\!:\!k]\right),
\qquad q\in\{T,R\}.
\label{eq:11}
\end{equation}
Here, $S_b^T=T_b$ and $S_b^R=R_b$, and $[1\!:\!k]$ denotes the corresponding $k$-candidate prefix. Both selections use the same demonstration target. We define the error ratio and preference label as
\begin{equation}
\begin{aligned}
\rho_b&=\frac{D_b^R+\epsilon}{D_b^T+\epsilon},\\
y_b&=\begin{cases}
+1,&\rho_b>1+\tau,\\
-1,&\rho_b<1-\tau,\\
0,&\text{otherwise}.
\end{cases}
\end{aligned}
\label{eq:12}
\end{equation}
The preference constructs a target from the detached pre-update distribution $P_b^{-}$ through normalized multiplicative reweighting~\cite{kivinen1997exponentiated}:
\begin{equation}
\begin{aligned}
P_b^*(c)&\propto P_b^{-}(c)\\
&\quad\cdot\exp\!\left[\log(1+\eta)y_b
\left(\mathbf 1_{c\in T_b}-\mathbf 1_{c\in R_b}\right)\right].
\end{aligned}
\label{eq:13}
\end{equation}
Before normalization, the preferred and disfavored sets are multiplied by $1+\eta$ and $(1+\eta)^{-1}$, respectively. Router minimizes the Kullback--Leibler (KL) divergence~\cite{cover2006elements} from the normalized target:
\begin{equation}
\mathcal L_R=\frac{1}{|\mathcal V|}
\sum_{b\in\mathcal V}
D_{\mathrm{KL}}\!\left(P_b^*\Vert P_{\theta,b}\right),
\label{eq:14}
\end{equation}
where $\mathcal V$ contains valid rated samples. Action errors determine the probability targets without backpropagating through discrete Top-$K$. A separate KL bound limits distribution changes, as in trust-region optimization~\cite{schulman2015trpo}. The preference tolerance, update magnitude, and constraints are specified in the supplementary material.

The hidden action representation conditions selection, while action prediction errors provide feedback for allocating visual precision and spatial coverage within the budget. Once Router training is complete, we freeze Router and continue training Fine for 2E using only its Top-\(K\) observations. This adapts action correction to the learned supplementary sets. Fine is trained for 3E in total, while Base remains frozen throughout.

\FloatBarrier
\section{Experiments}
\subsection{Action-relevant visual allocation and supplementation}
We evaluate whether the supplementary path in Section 3 improves closed-loop control over Base. On LIBERO~\cite{liu2023libero}, with seed 2 and 500 episodes per suite, fixed capacity $K=24$ raises the four-suite mean from $80.50\%$ for matched Base64 to $83.80\%$. All four suites improve in this seed-2 comparison. The archived Candidate512All run reaches $84.15\%$ with all 672 candidates.

\begin{figure}[!htbp]
\centering
\includegraphics[width=0.92\linewidth,height=0.40\textheight,keepaspectratio]{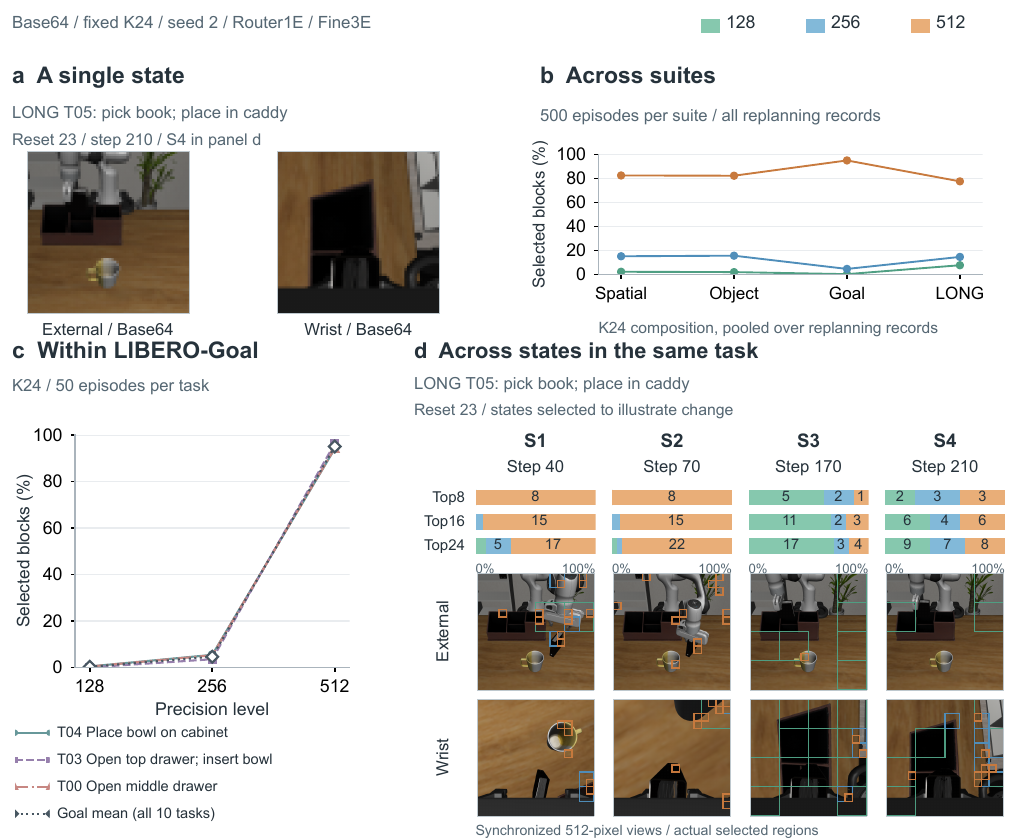}
\caption{Visual allocation with fixed $K24$ (seed 2, Router1E, Fine3E). (a) Base64 inputs at S4. (b) Suite-level precision counts. (c) Three Goal tasks and the suite mean. (d) LONG T05, reset 23, steps 40/70/170/210. Green/blue/orange denote 128/256/512. Bars show same-ranking Top8/16/24 proportions; labels give counts. Prefixes are not independently trained models.}
\label{fig:allocation}
\end{figure}

We next examine their precision composition. Under the same 24-patch budget, Goal concentrates more selections at precision 512, whereas LONG retains more lower-precision observations (Fig.~\ref{fig:allocation}(b)). The Goal tasks in Fig.~\ref{fig:allocation}(c) show a similar preference for precision 512. All patches have the same encoder input size. Different precision levels therefore imply different local sampling and coverage. These statistics describe the suites' local observation configurations, not changes in additional count.

To examine the relation between candidate ranking and additional count, we measure the precision composition of the first 8, first 16, and all 24 candidates from the same fixed-$K24$ model at the same replanning states. The three sets share one ranking and differ only in retained count. Figure~\ref{fig:allocation}(d) shows that the leading candidates favor precision 512 in S1 and S2, while lower-precision observations become more prominent in S3. In S4, the selected prefixes contain a mixture of all three precision levels. These changes illustrate how precision composition varies with both execution state and ranking cutoff. We next compare closed-loop performance across fixed capacities.

\FloatBarrier
\Needspace{8\baselineskip}
\subsection{Additional capacity and task differences}
We compare fixed capacities to test whether additional observations improve control consistently and whether tasks favor the same capacity. With Base and the candidate precision set fixed, we train separate $K\in\{8,16,24,32\}$ configurations. Each uses the same three-stage procedure: 1E of Router update opportunities and 3E of total Fine training. All results use seed 2 and 500 episodes per suite.

\begin{figure}[!htbp]
\centering
\includegraphics[width=0.92\linewidth,height=0.32\textheight,keepaspectratio]{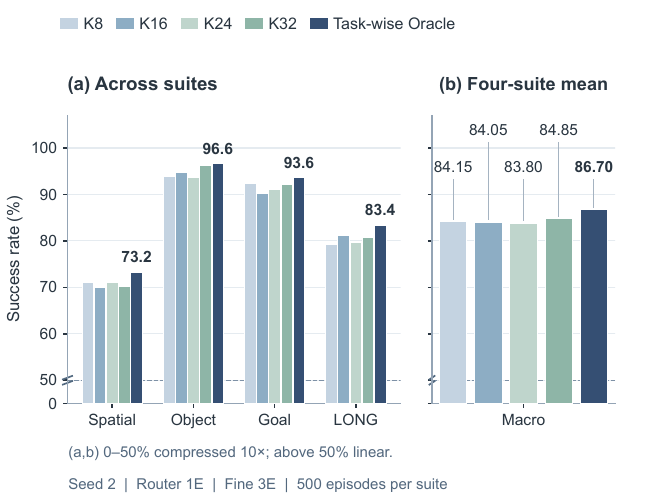}
\caption{Capacity comparison (seed 2, Router1E, Fine3E; 500 episodes per suite). (a) Suite success. (b) Four-suite mean. The task-wise Oracle is an analytical reference, not online selection. In both panels, 0--50\% is compressed tenfold; above 50\% is linear.}
\label{fig:capacity}
\end{figure}

As shown in Fig.~\ref{fig:capacity}(a,b), closed-loop success does not increase monotonically with additional count. Increasing $K$ from 8 to 32 quadruples the number of local observations but raises the four-suite mean by 0.70 percentage points. Intermediate budgets do not yield continuous gains. Although $K32$ has the highest overall mean, Goal and LONG attain their best suite results at $K8$ and $K16$, respectively. Under this training and evaluation setup, more observations provide limited average gains, and the best overall budget is not best for every suite.

Table~\ref{tab:best_k} reports the best tested $K$ for each task, with ties resolved toward smaller $K$. All Spatial tasks select at most $K24$, while Goal tasks T08 and T09 select $K32$ despite $K8$ achieving the highest Goal suite mean. Thus, tasks within the same suite can favor different capacities, and suite averages do not determine each task's preferred budget.

\begin{table}[!htbp]
\centering
\caption{Best $K$ per task among $K\in\{8,16,24,32\}$ (seed 2, Router1E, Fine3E; 50 episodes per task). Ties favor smaller $K$. Task IDs are suite-specific. This is a post-hoc comparison, not online budget selection.}
\label{tab:best_k}
\small
\renewcommand{\arraystretch}{1.03}
\setlength{\tabcolsep}{7pt}
\begin{tabular}{lrrrr}
\toprule
Task & Spatial & Object & Goal & LONG \\
\midrule
T00 & 24 & 8 & 16 & 8 \\
T01 & 8 & 16 & 8 & 8 \\
T02 & 24 & 24 & 8 & 8 \\
T03 & 8 & 32 & 8 & 32 \\
T04 & 16 & 32 & 16 & 24 \\
T05 & 24 & 16 & 8 & 8 \\
T06 & 8 & 32 & 8 & 32 \\
T07 & 24 & 16 & 8 & 16 \\
T08 & 8 & 8 & 32 & 16 \\
T09 & 24 & 8 & 32 & 8 \\
\bottomrule
\end{tabular}
\end{table}

Of the 40 tasks, 26 achieve their highest success among the tested budgets with at most 16 additional patches. Larger budgets offer no higher success on these tasks, although other tasks perform best with more patches. The benefit of additional capacity therefore varies by task.

Combining the task-level best configurations retrospectively gives a Task-wise Oracle mean of $86.70\%$, above the best uniform budget's $84.85\%$. This result reflects complementarity across budgets in the evaluated tasks. Because configuration selection and result aggregation use the same evaluation data, the Oracle is subject to selection bias~\cite{cawley2010overfitting}. It is an analytical reference, not the result of online capacity selection.

These results motivate combining action predictions at different capacities. The next section describes the construction and training of A32 multi-capacity action fusion. We compare it with fixed $K32$ at the same actual reading limit of 32 patches to test whether combining capacity-specific predictions improves closed-loop control.

\FloatBarrier
\subsection{Multi-capacity action fusion and closed-loop control}
Motivated by the complementary capacity configurations, A32 adds an action fusion head to the method in Section 3. It combines predictions from $\mathcal K=\{8,16,24,32\}$ under the same candidate ranking. All prefixes share one Top32 encoding. A softmax gate~\cite{jacobs1991adaptive} assigns learned budget weights to combine the corrected actions:
\begin{equation}
\begin{aligned}
\boldsymbol\omega_t&=\operatorname{softmax}(W_Av_t+b_A),\\
\hat{\mathbf a}_t^A&=\sum_{k\in\mathcal K}\omega_{t,k}\hat{\mathbf a}_t^{(k)}.
\end{aligned}
\label{eq:fusion}
\end{equation}
Here, $v_t$ summarizes the action context and routing statistics, and $\hat{\mathbf a}_t^{(k)}$ is the corrected prediction at budget $k$. For single-budget training, $\mathcal K=\{K\}$. For A32, $\mathcal K=\{8,16,24,32\}$, so Router feedback averages action errors over the four ranking prefixes. In the final stage, Base and Router remain frozen, while the fused-action L1 loss jointly trains Fine and the fusion head.

Across three training seeds, A32 achieves $84.23\pm1.44\%$ on LIBERO, improving over matched Base64 by 4.07 percentage points. It also outperforms the higher-resolution Base policies, including those trained for 30E (Table~\ref{tab:success}). In the matched seed-2 comparison, A32 improves over fixed $K32$ by 0.70 percentage points at the same 32-patch reading budget.

On RLBench10~\cite{james2020rlbench}, we use a single front camera, four Base visual sources, and 336 candidates. All comparisons use ResNet18, training seeds 2/3/4, and 100 evaluation episodes per task. A32 achieves $65.27\pm0.78\%$ mean success, exceeding Base64, Base128, and Base256 trained for 20E by 28.50, 23.07, and 14.97 percentage points, respectively. It also outperforms the Base128 and Base256 policies trained for 30E by 16.53 and 9.77 percentage points (Table~\ref{tab:success}).

\begin{table}[!htbp]
\centering
\caption{Closed-loop success (\%), mean $\pm$ sample standard deviation over training seeds 2/3/4. Each seed uses 500 episodes per LIBERO suite and 1,000 episodes on RLBench10. LIBERO Macro weights the four suites equally; its SD is computed across seed-level means.}
\label{tab:success}
\small
\renewcommand{\arraystretch}{1.00}
\setlength{\tabcolsep}{4pt}
\resizebox{\linewidth}{!}{%
\begin{tabular}{lrrrrrr}
\toprule
& \multicolumn{5}{c}{LIBERO} & RLBench10\\
\cmidrule(lr){2-6}\cmidrule(lr){7-7}
Configuration & Spatial & Object & Goal & LONG & Macro & Mean\\
\midrule
Base64 (20E) & 66.27 $\pm$ 0.31 & 92.53 $\pm$ 1.86 & 85.27 $\pm$ 3.93 & 76.60 $\pm$ 0.92 & 80.17 $\pm$ 0.29 & 36.77 $\pm$ 1.62\\
Base128 (20E) & 72.87 $\pm$ 3.58 & 92.00 $\pm$ 0.20 & 78.47 $\pm$ 6.00 & 72.73 $\pm$ 2.30 & 79.02 $\pm$ 2.48 & 42.20 $\pm$ 4.19\\
Base256$^\dagger$ (20E) & 63.20 $\pm$ 9.09 & 82.40 $\pm$ 6.81 & 67.00 $\pm$ 11.59 & 69.33 $\pm$ 6.54 & 70.48 $\pm$ 5.92 & 50.30 $\pm$ 2.34\\
Base128 (30E) & \textbf{73.73 $\pm$ 3.76} & 92.67 $\pm$ 2.44 & 76.40 $\pm$ 9.83 & 76.67 $\pm$ 1.10 & 79.87 $\pm$ 3.27 & 48.73 $\pm$ 2.75\\
Base256$^\dagger$ (30E) & 66.73 $\pm$ 6.52 & 86.33 $\pm$ 4.54 & 68.53 $\pm$ 11.38 & 69.93 $\pm$ 5.43 & 72.88 $\pm$ 5.27 & 55.50 $\pm$ 2.17\\
\midrule
A32 & 70.53 $\pm$ 1.89 & \textbf{96.13 $\pm$ 0.70} & \textbf{90.07 $\pm$ 1.29} & \textbf{80.20 $\pm$ 2.03} & \textbf{84.23 $\pm$ 1.44} & \textbf{65.27 $\pm$ 0.78}\\
\bottomrule
\end{tabular}}
\par\smallskip
{\footnotesize $^\dagger$Base256 uses ResNet50 on LIBERO and ResNet18 on RLBench10; other Base backbones are ResNet18.
A32 uses frozen Base64 (20E), Router1E, Fine3E, and action fusion over $K8/16/24/32$, reading 32 High patches.}
\end{table}

\FloatBarrier
\Needspace{8\baselineskip}
\subsection{Computational cost}
Table~\ref{tab:cost} reports the computational cost per 50-step action-planning call. Compared with Base128/\allowbreak{}ResNet18 (20E), A32 increases computation from 4.138 to 6.562 GFLOPs and inference latency from 12.92 to 22.60 ms, while improving three-seed mean success rates by 5.21 and 23.07 percentage points on LIBERO and RLBench10, respectively. Even compared with Base128 trained for 30E, A32 maintains improvements of 4.36 and 16.54 percentage points, respectively. A32 shares a single Top32 visual encoding across four capacity-specific action predictions, achieving higher closed-loop success rates through additional visual computation.
\begin{table}[!htbp]
\centering
\caption{Computational cost of seed-2 checkpoints on the same RTX6000D (batch size 1, FP32).}
\label{tab:cost}
\small
\renewcommand{\arraystretch}{1.00}
\setlength{\tabcolsep}{5pt}
\resizebox{\linewidth}{!}{%
\begin{tabular}{llrrrr}
\toprule
Model & Training budget & Params (M) & GFLOPs & \shortstack{Peak allocated\\memory (MiB)} & Latency (ms) \\
\midrule
ACT64 / ResNet18 & 20E & 52.036 & 1.482 & 220.71 & 12.34 \\
ACT128 / ResNet18 & 20E & 52.036 & 4.138 & 222.85 & 12.92 \\
ACT128 / ResNet18 & 30E & 52.036 & 4.138 & 222.85 & 12.61 \\
ACT256 / ResNet50 & 20E & 65.111 & 26.937 & 278.53 & 18.26 \\
ACT256 / ResNet50 & 30E & 65.111 & 26.937 & 278.53 & 19.33 \\
\midrule
E $K8$ & Base20E+Fine3E & 64.764 & 2.843 & 272.70 & 18.65 \\
E $K16$ & Base20E+Fine3E & 64.764 & 4.040 & 272.70 & 19.25 \\
E $K24$ & Base20E+Fine3E & 64.764 & 5.237 & 272.70 & 21.15 \\
E $K32$ & Base20E+Fine3E & 64.764 & 6.434 & 272.70 & 20.94 \\
\textbf{Atop $8/16/24/32$ (A32)} & \textbf{Base20E+Fine3E} & \textbf{64.766} & \textbf{6.562} & \textbf{272.71} & \textbf{22.60} \\
\bottomrule
\end{tabular}}
\par\smallskip
\begin{minipage}{\textwidth}\footnotesize
Peak memory denotes PyTorch peak allocated memory. Latency includes preprocessing, cropping, transfer, model execution, and CPU action output; it excludes simulation, rendering, disk access, and weight loading. E and Atop additionally use Router1E. Measurements use the same four Spatial states, 20 warm-up calls, and 200 synchronized timed calls. GFLOPs count convolution and matrix multiplication, with two operations per multiply-add.
\end{minipage}
\end{table}

\FloatBarrier
\section{Discussion and conclusion}
We study visual precision and capacity beyond a fixed base observation. With the same sampled pixel count, dense observations of smaller regions retain finer local detail but reduce spatial coverage. Our experiments show that higher input resolution or more supplementary observations do not necessarily improve control, and that tasks favor different capacities. Action-conditioned location--precision allocation and Base-aligned visual residuals enable useful visual supplementation while keeping Base frozen. Multi-capacity action fusion further combines the benefits of different configurations.

This supplementary mechanism may also extend to pretrained VLAs. A32 adds approximately 12.73M parameters to ACT. If the branch retains a similar size, it would account for a smaller fraction of larger models such as SmolVLA~\cite{shukor2025smolvla}. However, visual token organization and action representations differ across models, so the actual computational overhead and control gains require validation.

The current method uses a fixed maximum additional capacity. Fusing action predictions across budgets does not reduce the number of observations actually read. Future work could determine the ranking cutoff from the action context and jointly optimize Base and the supplementary branch. For pretrained VLAs such as SmolVLA, decisions about location, precision, and capacity could also be incorporated into pretraining or post-training, allowing visual resource allocation to adapt more directly to current action needs.

\Needspace{9\baselineskip}
\section*{LLM Usage Statement}
ChatGPT and Codex assisted with material retrieval and checking, organization of code and experimental records, figure--text integration, English translation, and language polishing. The authors formulated the research question and core method and are responsible for experimental design, evidence review, and the final conclusions. These tools also assisted with syntax corrections during code checking.

{\small
\bibliographystyle{ieeenat_fullname}
\bibliography{references}
}
\end{document}